%% file: main.tex
\documentclass[11pt]{article}

\usepackage[margin=1in]{geometry}
\usepackage{times}
\usepackage{microtype}

\usepackage{amsmath}
\usepackage{amssymb}
\input{math_commands.tex}

\usepackage{graphicx}
\usepackage{subcaption}
\usepackage{wrapfig}
\usepackage{booktabs}
\usepackage{multirow}
\usepackage{float}
\usepackage{array}

\usepackage[numbers,sort&compress]{natbib}

\usepackage{xcolor}
\usepackage[colorlinks=true,
            linkcolor=blue,
            citecolor=blue,
            urlcolor=blue]{hyperref}
\usepackage{url}
\title{HHR: Hierarchical Hash Retrieval for Efficient LLM Generation}
 
\author{
Lianjun Liu$^{1}$,
Tiantian Zheng$^{1}$,
You Huang$^{2}$,
Weiqi Yan$^{2}$,\\
Mingte Qiu$^{1}$,
Huazhong Liu$^{1,3}$,
Xiaofeng Zhu$^{1}$,
Yunshan Zhong$^{1}$\thanks{Corresponding author: \texttt{yszhong01@gmail.com}}\\[0.7em]
$^{1}$Hainan University, Haikou, China\\
$^{2}$Xiamen University, Xiamen, China\\
$^{3}$Zhejiang Normal University, Jinhua, China
}

\date{}

\begin{document}

\maketitle

\begin{abstract}

Efficient long-context inference is essential for large language models (LLMs), yet it poses a severe computational bottleneck.
Hash-based retrieval offers an efficient alternative by encoding queries and keys into binary codes and using Hamming distance for key selection. 
However, this leads to a critical mismatch between Hamming distance and attention relevance. Query-Key logits depend jointly on directional similarity and feature magnitudes, whereas hash binarization discards magnitude information, causing both false-positive retrieval of low-logit keys and false-negative omission of high-logit keys.
To address these failures, we propose \textbf{Hierarchical Hash Retrieval (HHR)}, a coarse-to-fine framework that progressively improves retrieval accuracy through \textbf{Geometry-Aware Key Routing (GKR)} and \textbf{Learned Hash Projection (LHP)}. 
GKR learns a head-wise orthogonal transformation to redistribute feature magnitudes and derive more discriminative page-level logit bounds, enabling effective pruning of low-logit keys while preserving important candidates. 
LHP then learns a head-wise projection space that aligns Hamming distance with the true Query-Key relevance ranking for fine-grained retrieval. 
By combining GKR and LHP, HHR suppresses false positives and recovers false negatives, substantially improving the fidelity of hash-based sparse attention. 
Extensive experiments across diverse LLMs and benchmarks demonstrate that HHR achieves superior performance over existing methods.
For example, on LongBench, HHR improves the average score by 1.10 points and, at a context length of 128K, achieves up to a $3.30\times$ decoding speedup and a 2.83$\times$ end-to-end speedup for Llama-3.1-8B-Instruct. The code is publicly available at \url{https://github.com/lianjunl13-sudo/HHR}.


%
%
%
\end{abstract}

\input{paper/Introduction}

\input{paper/related_work}
\input{paper/method}

\input{paper/experiments}

\section{Limitations and Future Work}

Despite achieving the best performance in certain cases, HHR still exhibits a performance gap relative to oracle Top-K selection. Therefore, developing more effective strategies remains an important focus for future work. For example, a stronger learning objective could potentially yield a better transformation $R$ and projection $\mathbf{W}$. Moreover, HHR is orthogonal to other compression techniques such as KV quantization. The effect of combining HHR with them remains an interesting question but is beyond the scope of this paper, and we leave it for future work.

\section{Conclusion}

In this paper, we identify a fundamental mismatch in hash-based sparse attention, where binarization discards feature magnitudes essential for Query–Key logits, causing false positives from directionally similar but low-logit keys and false negatives from high-logit keys with divergent directions. To address this, we propose Hierarchical Hash Retrieval (HHR), a coarse-to-fine framework combining Geometry-Aware Key Routing (GKR) and Learned Hash Projection (LHP). GKR learns orthogonal transformations to redistribute magnitudes and derive tighter page-level logit bounds for early pruning of low-logit keys, while LHP learns a head-wise projection aligning Hamming distance with true relevance ranking to reduce false negatives. Together, they substantially improve the fidelity of hash-based sparse attention. Experiments across diverse LLMs and benchmarks confirm HHR’s effectiveness and efficiency.


\clearpage

\subsection*{Acknowledgments}
 
This work was supported by the National Natural Science Foundation of China (No. 62606133).

\subsection*{AI use statement}

In this work, we used generative AI tools to assist with language polishing, grammatical revision, as well as partial code development and debugging. Generative AI tools were not used to replace experimental execution, generate experimental data, or serve as a source of unverified research conclusions or literature evidence.
All AI-assisted text, code, and research suggestions were manually reviewed and verified by the authors. For textual content, we checked technical accuracy, consistency of presentation, and correspondence with the relevant literature. AI-assisted code was verified through actual execution and experimental testing. Research ideas and technical analyses suggested with the assistance of generative AI were independently assessed based on theoretical derivations, experimental results, and relevant prior work.
The authors take full responsibility for the final content of this work, including all text, code, technical claims, and other research materials produced or revised with the assistance of generative AI.

\subsection*{Reproducibility statement}

We support the reproducibility of our results by providing complete descriptions of the proposed method, implementation details, and experimental settings. The main algorithms and training/inference procedures are presented in the main text, while the models, datasets, hyperparameters, evaluation metrics, and detailed experimental configurations are described in the experimental section and appendix. Additional ablation studies and supplementary results are also provided in the appendix. The source code and necessary experimental configurations have been included in the supplementary material to facilitate reproduction of the results reported in this paper.

\bibliographystyle{plainnat}
\bibliography{iclr2027_conference}

\clearpage
\appendix 

\section{Configurations Details}
\label{app:Configurations Details}

Table~\ref{tab:model_configs} summarizes the configurations of the models used in our experiments. For inputs exceeding the maximum context length, we truncate from both the beginning and the end.
Table~\ref{tab:baseline_configs} summarizes the configurations of the baseline methods used in our experiments.
Table~\ref{tab:loss_coefficients} summarizes the hyperparameters used in our training objective. For Geometry-Aware Key Routing, we set the candidate budget to 30\%, the final Top-$K$ ratio to 1.5\%, and the hash length to 128 bits.

\begin{table}[ht]
\centering
\small
\caption{Configurations of the models used for evaluation.}
\label{tab:model_configs}
\begin{tabular}{c|c|c}
\toprule
\textbf{Model} & \textbf{Configs} & \textbf{Values} \\
\midrule

\multirow{5}{*}{Llama-3.1-8B-Instruct}
& \#Layer & 32 \\
& \#Attention Heads & 32 \\
& \#KV Heads & 8 \\
& Hidden Size & 4096 \\
& Max Context Length & 131072 \\
\midrule

\multirow{5}{*}{Mistral-7B-Instruct}
& \#Layer & 32 \\
& \#Attention Heads & 32 \\
& \#KV Heads & 8 \\
& Hidden Size & 4096 \\
& Max Context Length & 32768 \\
\midrule

\multirow{5}{*}{Qwen3-4B}
& \#Layer & 36 \\
& \#Attention Heads & 32 \\
& \#KV Heads & 8 \\
& Hidden Size & 2560 \\
& Max Context Length & 40960 \\
\bottomrule
\end{tabular}
\end{table}

\begin{table}[ht]
\centering
\small
\caption{Configurations of the evaluated baseline methods.}
\label{tab:baseline_configs}

\begin{tabular}{
    >{\centering\arraybackslash}m{2.2cm}
    >{\raggedright\arraybackslash}m{5.8cm}
}
\toprule
\textbf{Method}
& \multicolumn{1}{c}{\textbf{Settings}} \\
\midrule

StreamingLLM
& Streaming KV cache compression with $4$ attention sink tokens and a retained cache ratio of $1.5\%$. \\
\midrule

MagicPIG
& Hash-based sparse attention with $K=8$, $L=40$, $4$ sink tokens, and a local window size of $64$, resulting in an effective candidate ratio of approximately $1.5\%$, comparable to TopK-$1.5\%$. \\
\midrule

Loki
& PCA-based Top-$k$ attention with $32$ selected channels and a Top-$k$ ratio of $1.5\%$. \\
\midrule

HATA
& Hash-based Top-$k$ attention with trained hash weights and a hash length of $128$ bits. \\
\midrule

HHR (Ours)
& Hierarchical hash-based retrieval with geometry-aware key routing and learned hash projection, using $128$-bit hash codes. \\
\bottomrule
\end{tabular}
\end{table}

\begin{table}
\centering
\small
\setlength{\tabcolsep}{8pt}
\renewcommand{\arraystretch}{1.1}
\caption{Loss coefficients used in our method.}
\label{tab:loss_coefficients}
\begin{tabular}{ccc}
\toprule
Coefficient & Corresponding loss & Value \\
\midrule
$\lambda_{1}$ & $\mathcal{L}_{\mathrm{precision}}$ & $0.05$ \\
$\lambda_{2}$ & $\mathcal{L}_{\mathrm{align}}$     & $0.25$ \\
$\lambda_{3}$ & $\mathcal{L}_{\mathrm{orth}}$      & $0.02$ \\
$\lambda_{4}$ & $\mathcal{L}_{\mathrm{decor}}$     & $0.02$ \\
$\lambda_{5}$ & $\mathcal{L}_{\mathrm{bal}}$       & $0.02$ \\
\bottomrule
\end{tabular}
\end{table}

\section{LongBench-v2 Results}
\label{app:longbench_v2}

We further evaluate HHR on LongBench-v2 in Table~\ref{tab:longbench_v2}. 
HHR achieves competitive performance across all three models. 
On Llama-3.1-8B-Instruct, HHR obtains the highest Overall score of 30.39, outperforming Full Attention, Oracle, and HATA. 
On Mistral-7B-Instruct-v0.3, HHR achieves an Overall score of 29.07, substantially outperforming HATA while remaining competitive with Full Attention and Oracle. 
For Qwen3-4B, HHR achieves the highest Hard Avg. score of 15.96 and improves the Overall score over HATA from 14.20 to 14.89, demonstrating its effectiveness on challenging long-context tasks

\begin{table}[ht]
\centering
\small
\caption{Evaluation results on LongBench-v2.}
\label{tab:longbench_v2}
\begin{tabular}{c c c c c}
\toprule
\textbf{Model} & \textbf{Method} & \textbf{Easy Avg.} & \textbf{Hard Avg.} & \textbf{Overall} \\
\midrule

\multirow{4}{*}{Llama-3.1-8B-Instruct}
& Full Attention
& 31.74 & 28.67 & 29.84 \\
& TopK (Oracle)
& 29.47 & 28.89 & 29.11 \\
\cmidrule(lr){2-5}
& HATA
& 31.90 & 29.17 & 30.21 \\
& HHR (Ours)
& \textbf{32.25} & \textbf{29.24} & \textbf{30.39} \\

\midrule

\multirow{4}{*}{Mistral-7B-Instruct-v0.3}
& Full Attention
& 30.36 & 28.97 & 29.50 \\
& TopK (Oracle)
& 32.61 & 27.42 & 29.40 \\
\cmidrule(lr){2-5}
& HATA
& 28.14 & \textbf{27.69} & 27.86 \\
& HHR (Ours)
& \textbf{32.24} & 27.12 & \textbf{29.07} \\

\midrule

\multirow{4}{*}{Qwen3-4B}
& Full Attention
& 14.51 & 15.52 & 15.13 \\
& TopK (Oracle)
& 14.51 & 15.52 & 15.13 \\
\cmidrule(lr){2-5}
& HATA
& \textbf{13.36} & 14.72 & 14.20 \\
& HHR (Ours)
& 13.15 & \textbf{15.96} & \textbf{14.89} \\

\bottomrule
\end{tabular}
\end{table}

\section{More Ablation Study}
\label{app:More Ablation Study}

\begin{figure}[h]
    \centering
        \centering
        \includegraphics[width=0.7\linewidth]{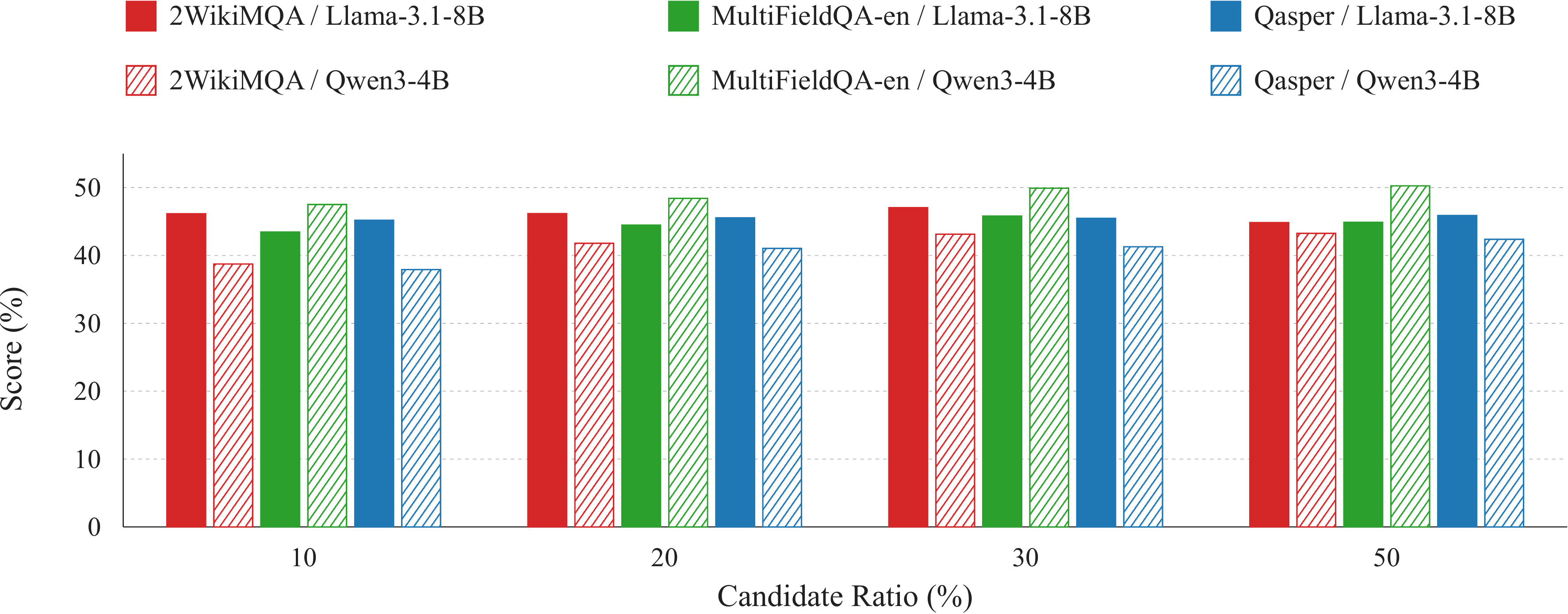}
        \caption{Comparison of model performance under different candidate ratios.}
        \label{fig:budget}
 
\end{figure}

\begin{figure}
    \centering
        \centering
        \includegraphics[width=0.7\linewidth]{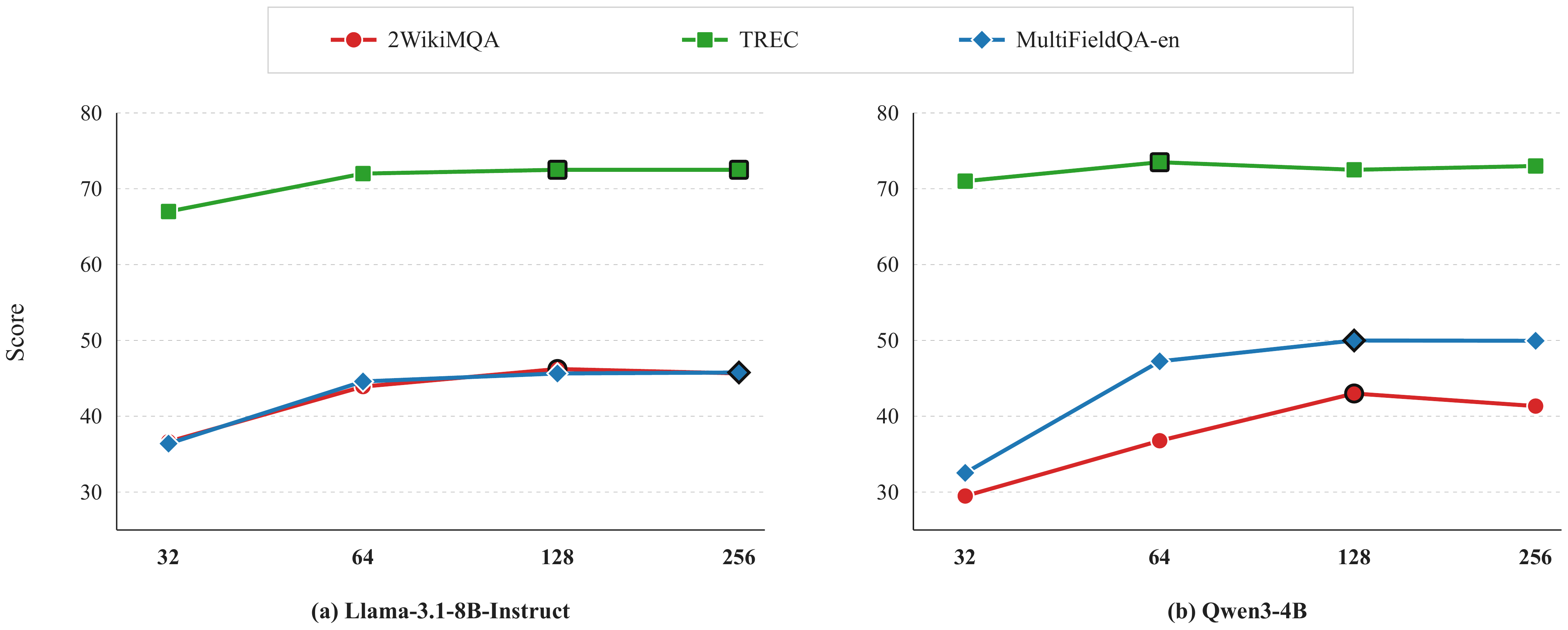}
        \caption{Sensitivity analysis of the number of hash bits.}
        \label{fig:hash_bit_ablation}

\end{figure}
 
\textbf{Budget on Geometry-Aware Key Routing.} As shown in Fig.~\ref{fig:budget}, different models exhibit varying sensitivity to the budget. For Llama-3.1-8B, budgets of 20\% and 30\% generally yield better overall performance, with the 30\% setting achieving the best results on 2WikiMQA and MultiFieldQA-en. In comparison, Qwen3-4B performs more consistently and achieves stronger overall performance with budgets of 30\% and 50\%.

\textbf{Hash Bit Length.}
We evaluate our HHR's sensitivity to the hash bit length using 32, 64, 128, and 256 bits in Fig.~\ref{fig:hash_bit_ablation}. Increasing the bit length from 32 to 128 significantly improves overall performance. However, further increasing it to 256 bits yields limited gains and even slightly degrades performance on some tasks. The 128-bit setting is the most stable across different models and datasets, and we therefore adopt it as the default setting.
 
\textbf{Effect of Loss Terms.} We conduct an ablation study on the loss terms in HHR, as shown in Table~\ref{tab:loss_ablation}. The full HHR model achieves the best performance on all datasets, with an average score of 44.36. Removing any single loss term leads to a consistent performance drop. Specifically, removing $\mathcal{L}_{\mathrm{align}}$ causes the largest degradation, reducing the average score to 42.83 (a drop of 1.53), which indicates that aligning the hash representations is crucial for effective retrieval. The absence of $\mathcal{L}_{\mathrm{precision}}$ also results in a notable decline to 43.35. In contrast, removing $\mathcal{L}_{\mathrm{orth}}$, $\mathcal{L}_{\mathrm{decor}}$, or $\mathcal{L}_{\mathrm{bal}}$ leads to smaller drops, with average scores of 43.59, 43.57, and 43.79, respectively. These results demonstrate that all loss terms contribute to the overall performance, while the alignment loss plays the most important role.

\begin{table}[t]
\centering
\small
\setlength{\tabcolsep}{5pt}
\renewcommand{\arraystretch}{1.05}
\caption{Ablation study of different loss terms in HHR.}
\label{tab:loss_ablation}
\begin{tabular}{lcccc}
\toprule
Setting & GovReport & HotpotQA & LCC & Avg. \\
\midrule
HHR (Ours) 
& \textbf{32.92} & \textbf{51.66} & \textbf{48.49} & \textbf{44.36} \\

w/o $\mathcal{L}_{\mathrm{precision}}$
& 32.06 & 50.51 & 47.48 & 43.35 \\

w/o $\mathcal{L}_{\mathrm{align}}$
& 31.71 & 49.87 & 46.91 & 42.83 \\

w/o $\mathcal{L}_{\mathrm{orth}}$
& 32.23 & 50.82 & 47.71 & 43.59 \\

w/o $\mathcal{L}_{\mathrm{decor}}$
& 32.26 & 50.78 & 47.67 & 43.57 \\

w/o $\mathcal{L}_{\mathrm{bal}}$
& 32.39 & 51.07 & 47.91 & 43.79 \\
\bottomrule
\end{tabular}
\end{table}

\textbf{Stage-wise Ablation Study.}
We further investigate the individual contributions of Geometry-Aware Key Routing (GKR) and Learned Hash Projection (LHP), with QUEST included as a retrieval baseline. As shown in Table~\ref{tab:stage_ablation}, using GKR alone improves over QUEST by 0.65, 0.57, and 0.63 points on QMSum, RepoBench-P, and SAMSum, respectively. LHP alone yields larger gains of 1.16, 1.15, and 1.14 points on the three datasets. When GKR and LHP are combined, HHR achieves 24.34, 47.62, and 43.38, outperforming QUEST by 1.82, 1.78, and 1.79 points, respectively. The full model consistently surpasses either individual module, indicating that GKR and LHP provide complementary benefits for improving retrieval quality.

\begin{table}[t]
\centering
\small
\setlength{\tabcolsep}{6pt}
\renewcommand{\arraystretch}{1.05}
\caption{Stage-wise ablation study of HHR.}
\label{tab:stage_ablation}
\begin{tabular}{lccc}
\toprule
Method & QMSum & RepoBench-P & SAMSum \\
\midrule
QUEST
& 22.52 & 45.84 & 41.59 \\

GKR
& 23.17 & 46.41 & 42.22 \\

LHP
& 23.68 & 46.99 & 42.73 \\

HHR
& \textbf{24.34} & \textbf{47.62} & \textbf{43.38} \\
\bottomrule
\end{tabular}
\end{table}
\end{document}

%% file: math_commands.tex
\usepackage{amsmath,amsfonts,bm}

\def\eqref#1{equation~\ref{#1}}

\def\1{\bm{1}}

\DeclareMathAlphabet{\mathsfit}{\encodingdefault}{\sfdefault}{m}{sl}
\SetMathAlphabet{\mathsfit}{bold}{\encodingdefault}{\sfdefault}{bx}{n}



%% file: paper/Introduction.tex
\section{Introduction}

The ability to process long contexts has become a critical requirement for large language models (LLMs)~\citep{Resource-efficient,LLM-Based} in a wide range of real-world applications, including document understanding, retrieval-augmented generation, and multi-turn reasoning over extended dialogues~\citep{Survey_Wang,Data_Foundations,llmsurvey}. 
However, the complexity of the self-attention mechanism with respect to sequence length imposes a severe computational bottleneck, making efficient long-context inference a pressing challenge~\citep{attention,LLM_Efficient_attention,Survey_Wang}. To mitigate this, sparse attention methods have been extensively explored \citep{wang2026fasa,zhu2026tactic,gao2026seerattention,gong2025hata}. These methods reduce the number of attended keys by selecting a small subset of important positions, thereby circumventing the need for full attention over the entire context~\citep{child2019generatinglongsequencessparse}.

Among the various sparse attention strategies~\citep{liu2026faster,choi2026retrospective}, hash-based retrieval has attracted considerable interest due to its adaptability and efficiency~\citep{desai2025hashattention,chen2025magicpig}. By projecting queries and keys into a compact space and encoding them into binary codes, hash-based methods first adaptively extract the important keys based on Hamming distance, which can be executed using efficient $\operatorname{XOR}$ and $\operatorname{POPCNT}$ operations, and then perform attention computation on these selected keys.
For instance, 
%
MagicPIG~\citep{chen2025magicpig} employs locality-sensitive hashing to select a small subset of candidate keys from the full KV cache.
%
%
HATA~\citep{gong2025hata} learns binary hash representations for queries and keys to approximate Query-Key similarity via Hamming distance.

Despite these gains, we uncover a critical mismatch in hash-based retrieval, where the magnitude information discarded for efficient hashing is essential for determining attention relevance. Specifically, Query-Key logits depend jointly on directional similarity and magnitudes, whereas hash binarization preserves only the signs of projected features and discards their magnitudes. 
As a result, Hamming distance deviates from the true Query-Key relevance, leading to inaccurate retrieval. As shown in Fig.~\ref{fig:motivation}, this mismatch gives rise to two complementary types of retrieval failure.
First, directionally similar keys yield small Hamming distances to the query yet low logits due to their low magnitudes, becoming false positives that waste budget in Region III. Second, keys with large magnitudes yield high logits to the query but diverge directionally, producing large Hamming distances and becoming false negatives that miss important tokens in Region II.

To address both failure modes, as shown in Fig.~\ref{fig:HHA}, we propose \textbf{Hierarchical Hash Retrieval (HHR)}, a novel coarse-to-fine framework that tackles false positives and false negatives. HHR comprises two complementary modules, including \textbf{Geometry-Aware Key Routing (GKR)} for coarse-grained pruning, and \textbf{Learned Hash Projection (LHP)} for fine-grained hash retrieval.
In the first stage, GKR partitions keys into non-overlapping pages and estimates the maximum logit of each page using per-dimension extrema. Pages with bounds below a threshold are discarded to eliminate low-relevance keys early. To obtain more discriminative bounds while preserving exact logits, GKR learns a head-wise orthogonal transformation that redistributes feature magnitudes across dimensions. The resulting extrema tighten the bounds for irrelevant pages, effectively pruning Region III false positives without sacrificing important keys.
In the second stage, LHP performs fine-grained hashing over the surviving candidates. Unlike conventional random projections, LHP learns a head-wise projection that aligns Hamming distance with the true relevance ranking. Differentiable sign relaxation enables gradient-based training, while orthogonality, bit-wise decorrelation, and balance regularization improve the quality of the binary space. Consequently, LHP substantially reduces Region II false negatives by assigning a more favourable hash space.

Extensive experiments across various LLMs and benchmarks demonstrate the effectiveness of the proposed HHR. For example, on LongBench, HHR achieves average scores of 46.96, 44.46, and 45.66 on Llama-3.1-8B-Instruct, Mistral-7B-Instruct, and Qwen3-4B, respectively, outperforming SOTA by 1.10, 2.82, and 2.69. At the same time, HHR achieves up to a 3.30$\times$ decode speedup and a 2.83$\times$ end-to-end speedup at a 128K context length on Llama-3.1-8B-Instruct.

%% file: paper/related_work.tex
\section{RELATED WORK}

\textbf{Structured sparse attention.} Sparse attention mitigates the inference cost of long‑context modeling by attending to only a subset of Query–Key pairs. Early approaches rely on structured sparsity patterns. For instance, LongGen combines local windows, attention sinks, and block‑sparse attention for efficient long‑context generation~\citep{ge2025longgen}. DuoAttention assigns full or restricted KV cache depending on each head's retrieval capability~\citep{xiao2025duoattention}. Star Attention employs distributed block‑sparse attention to scale to extremely long sequences~\citep{acharya2025star}. More recent methods move beyond fixed patterns by determining sparse regions in an input‑dependent manner. HiP progressively prunes irrelevant blocks through hierarchical search~\citep{lee2025hip}. 
\textbf{Dynamic sparse attention}. Dynamic sparse attention methods operate in an input-dependent manner~\citep{jiang2024minference,li2025mminference,choi2026retrospective,vasylenko2026sparse,desai2026vattention,zhu2026tactic,HISA}. FlexPrefill adaptively selects sparse patterns and budgets conditioned on queries and attention distributions~\citep{lai2025flexprefill}. SeerAttention introduces a learnable gating network to predict salient blocks~\citep{gao2024seerattention}. SpargeAttention discards low‑value computations via online block‑level filtering~\citep{zhang2025spargeattention}. Twilight adopts a hierarchical Top‑$p$ mechanism to dynamically adjust the sparsity budget~\citep{lin2025twilight}, suggesting that recent sparse attention is gradually shifting from predefined patterns toward query‑dependent selection. NSA and MoBA reduce the attention scope through hierarchical token selection and block‑level routing, respectively~\citep{yuan2025nsa,lu2025moba}. 
Another line of work leverages clustering or external memory to compress the search space, including Squeezed Attention, ClusterAttn, and Dynamic Block‑Sparse Attention~\citep{hooper2025squeezed,zhang2025clusterattn,xiao2025dynamicblock}. 
\textbf{Hash-based sparse attention.} Hash-based sparse attention represents another efficient paradigm that leverages hashing to identify important keys relevant to each query. HyperAttention~\citep{han2024hyperattention} employs locality-sensitive hashing (LSH) to identify dominant entries and approximates the remaining entries through sampling. MagicPIG~\citep{chen2025magicpig} leverages LSH to enable sampling-based attention approximation, using importance sampling over key-value pairs to estimate the attention output. HAX~\citep{zhan2025hax} combines LSH attention with Key Selection attention to capture content-dependent Query-Key interactions and shared attention to a small set of globally important keys, respectively. HashAttention~\citep{desai2025hashattention} and HATA~\citep{gong2025hata} both learn hash representations for queries and keys, using binary codes to approximate their similarity relationships for efficient pivotal-token selection. 
However, existing hash-based methods fail to consider the heterogeneous contributions of different feature dimensions to attention matching, leading to a misalignment between Hamming distance and the actual Query-Key relevance, making the retrieval accuracy unsatisfactory.

%% file: paper/method.tex
\begin{figure}
    \centering
    \includegraphics[width=0.9\linewidth]{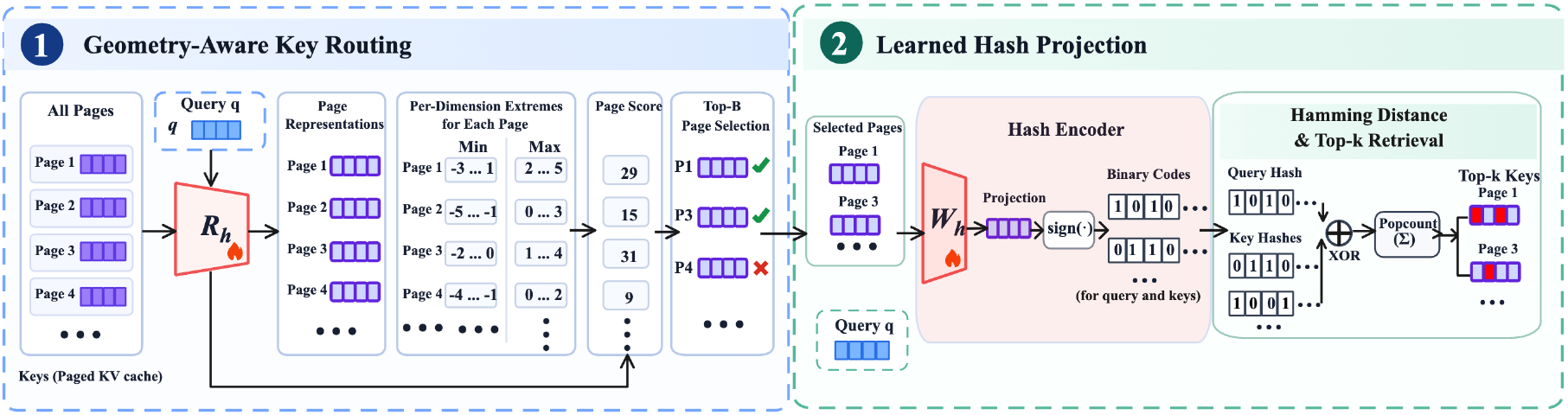}
    \caption{Overview of Hierarchical Hash Retrieva (HHR).}
    \label{fig:HHA}
    \vspace{-4mm}
\end{figure}

\section{Method}
\label{sec:method}

\subsection{Preliminary}

Hash-based retrieval methods~\citep{gong2025hata,desai2025hashattention,spotlight} project queries and keys into binary codes and use Hamming distance to estimate key importance. Given a hash function \(f(\cdot)\), the binary representations are obtained as:
\begin{equation}
b_q = \operatorname{sign}(f(q)), \quad b_k = \operatorname{sign}(f(k)),
\end{equation}
where \(\operatorname{sign}(\cdot)\) binarizes its input. The Hamming distance between two codes is then:
\begin{equation}
D_H(b_q,b_k)=\sum_{j=1}^{b} \mathbb{I}[b_{q,j}\neq b_{k,j}],
\end{equation} 
with \(b\) denoting the code length. This distance can be computed very efficiently via bitwise operations, offering substantial speedups over dense dot products.

\subsection{Motivation}
\label{sec:motivation}

Hash-based sparse attention methods typically use random projections to map queries and keys into a lower-dimensional hashing space, convert the projected features into binary codes, and retrieve keys according to Hamming distance~\citep{chen2025magicpig,desai2025hashattention}. However, Hamming distance can be poorly aligned with the actual Query-Key logits, leading to both false positive and false negative retrievals.

\begin{wrapfigure}{r}{0.5\linewidth}
    \centering
    \includegraphics[width=\linewidth]{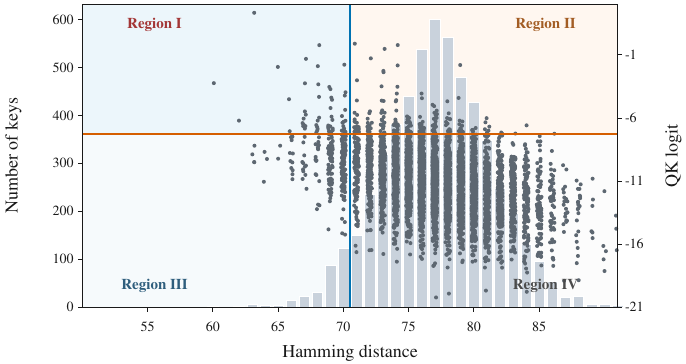}
    \caption{Distribution of Hamming distances obtained using a random orthogonal projection followed by $\operatorname{sign}(\cdot)$, together with the corresponding Query-Key logits.}
    \label{fig:motivation}
\end{wrapfigure}

To characterize this misalignment, Fig.~\ref{fig:motivation} compares the Hamming distance with the Query-Key logits. The orange dashed line denotes the Rop-$5\%$ boundary according to Query-Key logits, while the blue dashed line denotes the Top-$5\%$ boundary according to Hamming distance. Their intersection divides the key space into four regions:
\begin{itemize}
\item \textbf{Region I (upper-left):} Keys with small Hamming distances and high Query-Key logits, which are correctly identified as important by both hashing and original attention.
\item \textbf{Region II (upper-right):} Keys with large Hamming distances but high Query-Key logits, which are important for attention but are missed by hash retrieval.
\item \textbf{Region III (lower-left):} Keys with small Hamming distances but low Query-Key logits, which are incorrectly retained by hash retrieval.
\item \textbf{Region IV (lower-right):} Keys with large Hamming distances and low Query-Key logits, which are correctly identified as unimportant by both mechanisms.
\end{itemize}
Regions II and III stem from the misalignment between Hamming similarity and Query-Key relevance. Specifically, the Query-Key logit between a query $q$ and a key $k$ is $q^\top k = \|q\|\|k\|\cos\theta$, where $\cos\theta$ captures directional similarity and $\|q\|,\|k\|$ determine the magnitude. Hamming distance, in contrast, depends only on binary signs and discards magnitudes, so it mainly reflects directional consistency in hash space but cannot preserve the magnitude information that determines key importance in attention~\citep{ALSH}.
%

This misalignment produces two retrieval failures. First, keys with similar directions but different magnitudes from the query yield small Hamming distances but low Query-Key logits, causing false positives in Region III, where low-logit keys consume the limited retrieval budget. Second, keys with large magnitudes but directions differing from the query yield high Query-Key logits but large Hamming distances, causing false negatives in Region II, where high-logit keys are discarded and attention misses important tokens.


\subsection{Hierarchical Hash Attention}
\label{sec:HHA}

Motivated by the above analysis, we propose Hierarchical Hash Retrieval (HHR), which addresses the two types of retrieval failure via a coarse‑to‑fine strategy combining Geometry‑Aware Key Routing (GKR) and Learned Hash Projection (LHP), as illustrated in Fig.~\ref{fig:HHA}.
%


\subsubsection{Geometry-Aware Key Routing}
\label{sec:GKR}

To filter out low-logit keys early and save the retrieval budget for important keys, we introduce \textbf{Geometry-Aware Key Routing (GKR)} as a coarse-grained first stage.
GKR partitions each head's keys into non-overlapping pages, estimates an upper bound on the maximum logit of each page~\citep{tang2024quest}, and discards pages whose bounds fall below a threshold.
Thus, before fine-grained retrieval, pages dominated by low-logit keys (Region III) are discarded, while those potentially containing high-logit keys (Region II) are retained.

%
%

\begin{figure}
    \centering
    \includegraphics[width=0.9\linewidth]{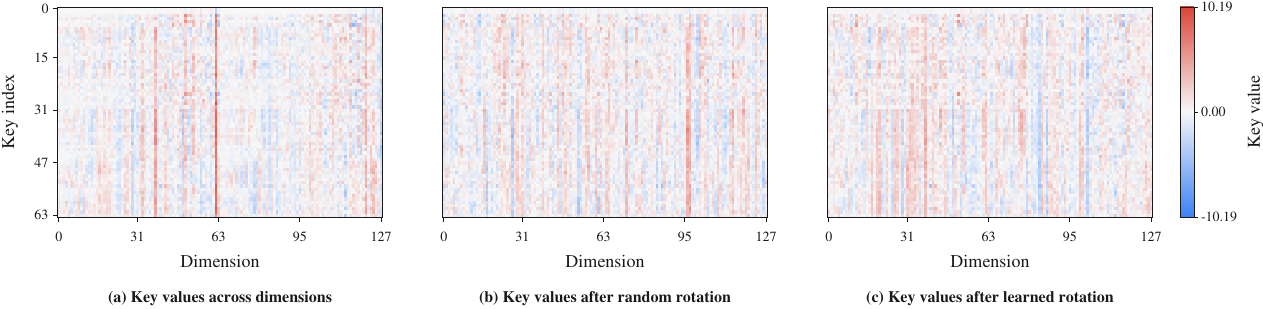}
    \caption{Distribution of keys before and after applying the learned transformation $R_h$. (a) Original key distribution across 128 dimensions for 64 consecutive keys. (b) The same keys after applying a random orthogonal rotation. (c) The same keys after applying the learned orthogonal matrix $R_h$.}
    \label{fig:rotation}
    \vspace{-4mm}
\end{figure}

For the $h$-th head and page $p$, let $q_h$ denote a query, and $l_{h,p}^j$ and $u_{h,p}^j$ denote the minimum and maximum of the keys in page $p$ along dimension $j$. The maximum contribution of dimension $j$ to the Query-Key logit is
$\max(q_h^j l_{h,p}^j,\; q_h^j u_{h,p}^j)$, so the page-level estimated upper bound is:
\begin{equation}
U_{h,p} = \sum\nolimits_{j=1}^{d} \max\left(q_h^j l_{h,p}^j,\; q_h^j u_{h,p}^j\right),
\label{eq:upperbound}
\end{equation}
where $d$ is the feature dimension. Since every key's value lies between the two extrema, $U_{h,p}$ bounds the largest logit attainable in the page.
Given the threshold $c_h$, GKR retains a page $p$ if $U_{h,p} > c_h$ and discards it otherwise, thereby removing pages unlikely to contain high-logit keys.

Importantly, bound quality depends on how keys distribute across dimensions. The same logits yield different dimension-wise extrema under different coordinate systems~\citep{KIVI,wang2026fasa,KVQuant}. An orthogonal transformation thus offers a degree of freedom: it changes the extrema while preserving the logits exactly. We therefore learn a head-wise orthogonal matrix $R_h$ applied to both query and key before computing the bound:
\begin{equation}
\tilde{q}_h = q_h R_h, \qquad \tilde{k}_{h,p,i} = k_{h,p,i} R_h,
\end{equation}
where $k_{h,p,i}$ is the key at position $i$ in page $p$ of head $h$. Since $R_h R_h^\top = I$, the transformation preserves the exact Query-Key logits: $\tilde{q}_h(\tilde{k}_{h,p,i})^\top
=
q_hR(k_{h,p,i}R_h)_\top
=
q_hR_h R_h^\top (k_{h,p,i})^\top
=
q_h (k_{h,p,i})^\top$.
Thus, a proper $R_h$ can leave attention unchanged but redistributes feature values, allowing dimension-wise extrema to yield more discriminative page-level bounds. GKR then evaluates Eq.~\ref{eq:upperbound} on the transformed keys, learning $R_h$ so that pages with important keys get large bounds, while irrelevant pages get tighter bounds.

%

To this end, GKR optimizes $R_h$ with three complementary objectives: (1) $\mathcal{L}_{\mathrm{recall}}$ keeps pages containing important keys, (2) $\mathcal{L}_{\mathrm{precision}}$ suppresses unnecessary page selection, and (3) $\mathcal{L}_{\mathrm{align}}$ aligns the retention score distribution with the relevance ranking of full attention.
In particular, during training, the exact logits provide an oracle. For head $h$, the Top-K keys with the largest logits define important pages, \emph{i.e.}, pages containing at least one such key. Let $M_h$ be the important pages currently filtered out ($U_{h,p} \le c_h$). For each missed page $p \in M_h$, its bound should exceed the threshold by a margin $m$:
\begin{equation}
\mathcal{L}_{\mathrm{recall}}
=
\frac{1}{|M_h|}
\sum\nolimits_{p \in M_h}
\max\left(0,\; c_h + m - U_{h,p}\right),
\end{equation}
with $\mathcal{L}_{\mathrm{recall}} = 0$ when $M_h$ is empty. This objective encourages missed pages to exceed the threshold by a safety margin, thereby preventing high‑logit keys from being discarded.
High recall alone over-selects pages, so we add a precision objective to suppress unnecessary page selection. 
Let $F_h$ be the selected pages containing no oracle Top-K key (false positives), and let $U_{h,p}^{*}$ be the oracle upper bound of page $p$, \emph{i.e.}, its largest exact Query-Key logit. The gap $U_{h,p} - U_{h,p}^{*}$ measures the overestimation of the estimated upper bound, which is quantified by:
$
s_{h,p} = \min(\frac{1}{\sigma_h}\max(0, U_{h,p}-U_{h,p}^{*}),\, s_{\max})
$,
where $\sigma_h$ is the standard deviation of $\{U_{h,p}^{*}\}$ over the pages of head $h$ and $s_{\max}$ is the maximum allowed overestimation. The precision loss is then defined as:
\begin{equation}
\mathcal{L}_{\mathrm{precision}}
=
\frac{
\sum_{p \in F_h} (1 + s_{h,p}) \log(1 + s_{h,p})
}{
\sum_{p \in F_h} (1 + s_{h,p})
},
\end{equation}
with $\mathcal{L}_{\mathrm{precision}} = 0$ when $F_h$ is empty. 
This objective penalizes false-positive pages with overestimated bounds, encouraging the learned transformation to tighten those bounds and reduce unnecessary selections.
%
%
Then, to retain overall ranking, we add a distribution-level objective that couples the estimated upper bound with the hash-based score.
Let $D_{h,i}$ be the pre-binarization distance~\citep{li2021harmonization} between the query and key $i$ under head $h$. Aggregating token-level distances with log-sum-exp gives the page-level Hash score:
\begin{equation}
H_{h,p} = \log \sum\nolimits_{i \in p} \exp\left(-D_{h,i} / \tau_h\right),
\end{equation}
where $\tau_h$ is the temperature. 
The negative sign turns distance into similarity, so a larger $H_{h,p}$ indicates a more relevant page. 
After normalization of $\operatorname{zscore}(x) = (x - \mu_x)/\sigma_x$, we combine the estimated upper bound and the page-level Hash score into a single retention score $C_{h,p} = \operatorname{zscore}(U_{h,p}) + \operatorname{zscore}(H_{h,p})$, and align it with the page relevance distribution of full attention:
\begin{equation}
P_{h,p}^{*} = \operatorname{softmax}_p\left(\frac{\operatorname{zscore}(U_{h,p}^{*})}{\tau_{\mathrm{page}}}\right),
\qquad
\mathcal{L}_{\mathrm{align}} = \operatorname{KL}\left(P_h^{*} \,\middle\|\, \operatorname{softmax}_p(C_h)\right),
\end{equation}
where $\tau_{\mathrm{page}}$ is the temperature. 
Unlike recall and precision, which affect only the estimated upper bound, $\mathcal{L}{\mathrm{align}}$ encourages the combined retention score to both match the page relevance distribution induced by the exact Query-Key logits, thereby coupling the estimated upper bound with the Hash-based score to preserve the relevance ranking of full attention.
Finally, the three objectives are combined as:
\begin{equation}
\mathcal{L}_{R_h} = \mathcal{L}_{\mathrm{recall}} + \lambda_1 \mathcal{L}_{\mathrm{precision}} + \lambda_2 \mathcal{L}_{\mathrm{align}},
\label{eq:overall-loss}
\end{equation}
where $\lambda_1$ and $\lambda_2$ control the contributions of the precision and distribution alignment objectives, respectively.
To keep $R_h$ orthogonal during optimization, we parameterize it with the Cayley transform~\citep{Helfrich}: $R_h
=
(I+A)^{-1}(I-A)$, where $A$ is skew-symmetric, $A^\top = -A$. Since $I + A$ is invertible, $R_h$ is orthogonal by construction, so optimizing $A$ adapts the feature distribution to Eq.~\ref{eq:overall-loss} while preserving the exact Query-Key logits.


Fig.~\ref{fig:rotation} shows that original keys are dominated by outlier dimensions~\citep{MUSTAFAR,KVQuant}. Random rotation partially mitigates this, but the learned transformation yields a more favorable, less dimension-dominated distribution with tighter, more discriminative bounds. Thus, the page-level upper bound separates relevant from irrelevant pages effectively. For example, at page size 32, retaining 10\% of candidates and treating Top-5\% keys as important, GKR improves Top-5\% recall from 43.20\% to 47.48\%. Overall, GKR learns an orthogonal transformation that preserves Query-Key logits while reorganizing the feature space, yielding discriminative bounds that prune Region III pages while achieving a better recall–budget trade-off and substantially improving the retention of high logit keys for Region II keys.


\subsubsection{Learned Hash Projection}
\label{sec:LHP}

After the first stage, a much smaller candidate set remains for fine-grained retrieval. We therefore introduce \textbf{Learned Hash Projection (LHP)} as the second fine-grained stage, which learns a compact projection space aligning hash distance with Query-Key relevance, so binary codes better preserve directional similarity and support efficient retrieval.
\begin{wrapfigure}{r}{0.58\linewidth}
    \centering
    \includegraphics[width=\linewidth]{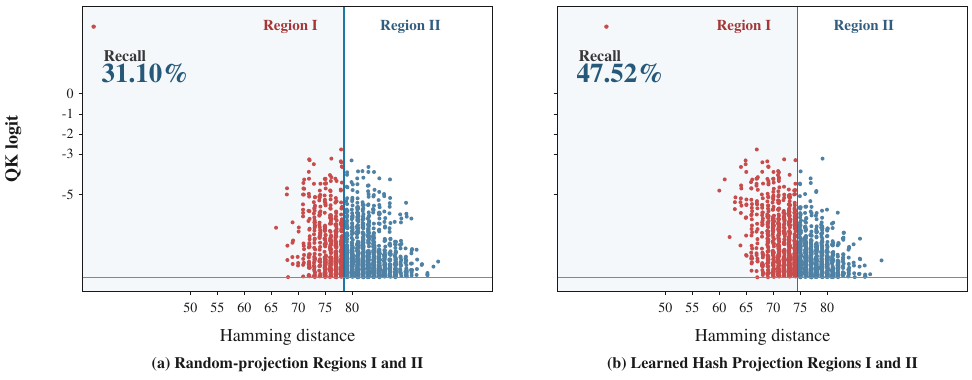}
    \caption{Top-10\% key recall from layer 25, KV head 4 of Llama-3.1-8B-Instruct.}
    \label{fig:hash}
     
\end{wrapfigure}


Specifically, for head $h$, we learn a projection matrix $\mathbf{W}_{h}\in\mathbb{R}^{d\times b}$, where $b$ is the hash code length. $\mathbf{W}_h$ is trained offline on keys retained from GKR-selected pages. Since $\operatorname{sign}(\cdot)$ is non-differentiable, we use a differentiable relaxation during training. For input $\mathbf{x}$, its projected features and relaxed hash code are defined as:
%
\begin{equation}
\mathbf{z} = \mathbf{x}\mathbf{W}_h, \qquad 
\widehat{\mathbf{c}} = \tanh\left(0.1\,\mathbf{z}\right),
\label{eq:soft_hash}
\end{equation}
where the scaling factor $0.1$ controls the saturation behavior of the $\tanh$ nonlinearity.
During training, we compute the soft hash distance $D_i$ between the query and the $i$-th candidate key $\mathbf{k}_{h,i}$ using these relaxed hash codes from Eq.~\ref{eq:soft_hash}. At inference time, the relaxed codes are replaced by their binary counterparts $\mathbf{c} = \operatorname{sign}(\mathbf{z})$, allowing for efficient Hamming-distance-based retrieval.

To train $\mathbf{W}_h$, we first define a soft selection probability $m_i = \sigma\!\left(\frac{D_{N+1}-D_i}{\tau_{\mathrm{sel}}}\right)$ for each candidate key $\mathbf{k}_{h,i}$, where $\sigma(\cdot)$ is the sigmoid function, $D_{N+1}$ is the hash distance of the $(N+1)$-th closest candidate, and $\tau_{\mathrm{sel}}$ is a temperature parameter controlling the sharpness of the soft selection. A smaller $D_i$ thus yields a larger selection probability $m_i$.
Let $\mathcal{K}_h^{\star}$ denote the set of Top-N relevant keys within the candidate set retained by GKR, determined according to their exact attention logits. We formulate the retrieval objective as:
\begin{equation}
\mathcal{L}_{\mathrm{retrieval}} = -\log\bigl(\frac{1}{N}\sum\nolimits_{i\in\mathcal{K}_h^{\star}} m_i\bigr).
\label{eq:retrieval_loss}
\end{equation}
Minimizing this objective encourages highly relevant keys to obtain smaller hash distances, thereby increasing their probability of being retrieved in the fine-grained hashing stage.
In addition to the retrieval objective, we introduce three regularization terms. First, we enforce an orthogonality constraint on the projection matrix:
\begin{equation}
\mathcal{L}_{\mathrm{orth}} = \left\|\mathbf{W}_h^{\top}\mathbf{W}_h-\mathbf{I}\right\|_{F}^{2},
\label{eq:orth_loss}
\end{equation}
which encourages different projection dimensions to capture complementary directions in the input space.
Second, given a training batch of $M$ relaxed hash codes $\widehat{\mathbf{c}}$, we stack them into $\widehat{\mathbf{C}}\in\mathbb{R}^{M\times b}$, center them along the sample dimension as $\boldsymbol{\mu}=\tfrac{1}{M}\sum\nolimits_{i=1}^{M}\widehat{\mathbf{c}}_i, \widetilde{\mathbf{C}}=\widehat{\mathbf{C}}-\mathbf{1}\boldsymbol{\mu}^{\top}$, and compute the bit-wise covariance matrix by $\boldsymbol{\Sigma} = \frac{1}{M}\widetilde{\mathbf{C}}^{\top}\widetilde{\mathbf{C}}$. We then define the decorrelation loss as:
\begin{equation}
\quad \mathcal{L}_{\mathrm{decor}} = \left\|\boldsymbol{\Sigma} - \operatorname{Diag}(\boldsymbol{\Sigma})\right\|_{F}^{2},
\label{eq:decor_loss}
\end{equation}
which suppresses correlations among different hash bits and thus reduces redundancy in the binary representations.
Third, to prevent individual hash bits from collapsing to predominantly positive or negative values, we introduce a bit-balance loss:
\begin{equation}
\mathcal{L}_{\mathrm{bal}}
= \bigl\lVert \frac{1}{M}\sum\nolimits_{i=1}^{M}\widehat{\mathbf{c}}_i \bigr\rVert_{2}^{2},
\label{eq:balance_loss}
\end{equation}
which encourages each bit to maintain balanced positive and negative activations across training samples, thereby avoiding degenerate codes and improving the effective utilization of the binary code space.
The overall LHP training objective is obtained by combining the above terms:
\begin{equation}
\mathcal{L}_{\mathrm{LHP}} = \mathcal{L}_{\mathrm{retrieval}} + \lambda_{3}\mathcal{L}_{\mathrm{orth}} + \lambda_{4}\mathcal{L}_{\mathrm{decor}} + \lambda_{5}\mathcal{L}_{\mathrm{bal}},
\label{eq:total_loss}
\end{equation}
where $\lambda_{3}$, $\lambda_{4}$, and $\lambda_{5}$ balance the objectives.
Unlike HATA~\citep{gong2025hata}, LHP only identifies the Top-N relevant keys and ignores the rest, avoiding noisy low-relevance keys. As shown in Fig.~\ref{fig:hash}, for Top-10\% key selection, the learned projection raises key recall from 31.10\% to 47.52\%, reducing misalignment between Hamming-distance retrieval and Query-Key logit ranking. Overall, LHP learns a compact projection space where hash distance better aligns with Query-Key relevance, enabling more accurate binary retrieval that transfers candidates from Region II to Region I.


%% file: paper/experiments.tex
\section{Experiments}
\label{sec:experiments}

\subsection{Experimental Setup}
\label{sec:experimental_setup}


\textbf{Baselines and configurations.} 
We compare HHR against seven baselines. Full Attention serves as the dense reference. Under the same token budget, Oracle selects the exact Top-$K$ keys based on real Query-Key attention scores. The sparse retrieval baselines include HATA~\citep{gong2025hata}, MagicPIG~\citep{chen2025magicpig}, and Loki~\citep{singhania2024loki}. For KV cache compression, StreamingLLM~\citep{xiao2024efficient} retains attention sinks and recent tokens, whereas SnapKV~\citep{li2024snapkv} selects informative cached positions from an observation window. Each baseline follows its recommended configuration under a comparable token budget. All methods use the Top-K ratio of 1.5\%, the same prompt templates, and retain full attention in the first two layers. Additional experimental details and the hyperparameter settings are provided in Appendix~\ref{app:Configurations Details}.
All experiments are conducted on a single NVIDIA A800 GPU with PyTorch~\citep{ansel2024pytorch}.

\textbf{Models, datasets, and metrics.} HHR is evaluated on Llama-3.1-8B-Instruct~\citep{grattafiori2024llama}, Mistral-7B-Instruct-v0.3~\citep{DBLP:journals/corr/abs-2310-06825}, and Qwen3-4B~\citep{yang2025qwen3} by using LongBench~\citep{bai2024longbench} and RULER~\citep{hsieh2024ruler} datasets.
LongBench contains 16 English datasets covering question answering, summarization, few-shot learning, synthetic retrieval, and code completion, on which we report our results. Evaluation follows the official metric for each dataset, including F1, ROUGE-L, accuracy, and code similarity. RULER covers 11 tasks at a context length of 32K. These tasks examine retrieval, aggregation, and question answering. Both task-level and average accuracy are reported.


\subsection{Main Results}
\label{sec:accuracy_evaluation}

\textbf{Results on LongBench.} As shown in Table~\ref{tab:longbench}, on LongBench, HHR demonstrates strong performance across all three model families. Compared with other sparse-attention and KV cache compression methods, HHR achieves average scores of 46.96, 44.46, and 45.66 on Llama-3.1-8B-Instruct, Mistral-7B-Instruct, and Qwen3-4B, respectively, obtaining the best results among these methods for each model. HHR also maintains consistent advantages on single-document QA, summarization, and few-shot tasks. These results indicate that HHR can effectively preserve task-relevant information and achieve strong cross-task generalization.

\begin{table*}[t]
\centering
\caption{Evaluation results on LongBench.}
\label{tab:longbench}
\setlength{\tabcolsep}{3.2pt}
\vspace{-2mm}
\scriptsize
\begin{tabular}{llccccccc}
\toprule
\multirow{2}{*}{Model}
& \multirow{2}{*}{Method}
& Single-Doc
& Multi-Doc
& \multirow{2}{*}{Summarization}
& \multirow{2}{*}{Few-shot}
& \multirow{2}{*}{Synthetic}
& \multirow{2}{*}{Code}
& \multirow{2}{*}{Mean} \\
& & QA & QA & & & & & \\
\midrule

\multirow{8}{*}{\shortstack[l]{Llama-\\3.1-8B-\\Instruct}}
& Full Attention
& 40.56 & 44.37 & 29.18 & 69.28 & 54.19 & 50.80 & 47.51 \\
& TopK (Oracle)
& 40.93 & 44.33 & 28.98 & 69.37 & 53.53 & 50.59 & 47.44 \\
\cline{2-9}
& MagicPIG
& 38.04 & 43.55 & 27.46 & 68.09 & 53.37 & 46.94 & 45.75 \\
& SnapKV
& 31.14 & 42.49 & 20.23 & 58.94 & \textbf{54.17} & 48.30 & 41.46 \\
& StreamingLLM
& 14.95 & 10.37 & 13.61 & 32.96 & 1.25 & 48.98 & 19.76 \\
& Loki
& 31.84 & 34.19 & 27.26 & 62.83 & 42.62 & \textbf{52.11} & 41.11 \\
& HATA
& 40.52 & 43.74 & 26.64 & 68.83 & 53.56 & 43.75 & 45.86 \\
& HHR (Ours)
& \textbf{40.63} & \textbf{44.29} & \textbf{28.39}
& \textbf{69.32} & 54.16 & 47.55 & \textbf{46.96} \\

\midrule

\multirow{8}{*}{\shortstack[l]{Mistral-7B-\\Instruct}}
& Full Attention
& 36.76 & 38.14 & 28.90 & 70.75 & 51.00 & 52.85 & 45.71 \\
& TopK (Oracle)
& 35.98 & 37.63 & 28.62 & 70.96 & 50.75 & 52.48 & 45.38 \\
\cline{2-9}
& MagicPIG
& 30.20 & 32.08 & 26.18 & 66.12 & 41.04 & 50.85 & 40.47 \\
& SnapKV
& 27.22 & 32.00 & 20.07 & 58.83 & \textbf{49.50} & 50.42 & 38.39 \\
& StreamingLLM
& 12.59 & 12.78 & 15.28 & 34.66 & 3.50 & 45.46 & 20.24 \\
& Loki
& 30.08 & 28.02 & 25.50 & 64.54 & 30.06 & \textbf{51.51} & 37.97 \\
& HATA
& 32.21 & 34.45 & 25.14 & 68.82 & 46.27 & 45.94 & 41.64 \\
& HHR (Ours)
& \textbf{34.85} & \textbf{37.00} & \textbf{28.06}
& \textbf{70.39} & \textbf{49.50} & 50.72 & \textbf{44.46} \\

\midrule

\multirow{8}{*}{\shortstack[l]{Qwen3-\\4B}}
& Full Attention
& 40.04 & 43.43 & 27.42 & 68.46 & 48.62 & 51.16 & 46.10 \\
& TopK (Oracle)
& 39.57 & 43.91 & 27.46 & 68.44 & 48.38 & 50.36 & 45.98 \\
\cline{2-9}
& MagicPIG
& 35.15 & 40.01 & 23.81 & 64.89 & 44.25 & 49.27 & 42.41 \\
& SnapKV
& 30.84 & 40.17 & 17.70 & 58.56 & \textbf{48.50}
& \textbf{50.40} & 39.98 \\
& StreamingLLM
& 11.65 & 16.49 & 14.98 & 25.81 & 1.75 & 47.66 & 19.10 \\
& Loki
& 35.49 & 40.34 & 26.65 & 63.92 & 47.75 & 46.14 & 42.94 \\
& HATA
& 37.19 & 42.84 & 25.15 & 66.53 & 48.34 & 37.84 & 42.97 \\
& HHR (Ours)
& \textbf{39.97} & \textbf{43.66} & \textbf{28.66}
& \textbf{68.59} & 48.38 & 45.58 & \textbf{45.66} \\

\bottomrule
\end{tabular}
\vspace{-3mm}
\end{table*}
%
%








\textbf{Results on RULER.} As shown in Table~\ref{tab:ruler}, on RULER, HHR also consistently outperforms other methods for all three model families. HHR maintains accuracy close to Full Attention. In contrast, SnapKV, MagicPIG, and Loki suffer substantial performance degradation. For example, HHR achieves a mean performance of 76.23 and 82.71 on Mistral-7B-Instruct and Qwen3-4B, improving the SOTA by 6.43 and 4.02, respectively.

\begin{table*}
\centering
\caption{
Evaluation results on RULER with a context length of 32K.
S1--S3 denote NIAH Single 1--3;
MK1--MK2 denote NIAH Multi-key 1--2;
MV and MQ denote NIAH Multi-value and NIAH Multi-query;
CWE and FWE denote Common Words Extraction and Frequent Words Extraction;
QA1 and QA2 denote question answering on SQuAD and HotpotQA, respectively.
}
\label{tab:ruler}
\vspace{-2mm}
\scriptsize
\resizebox{\textwidth}{!}{
\begin{tabular}{llcccccccccccc}
\toprule
Model & Method &
S1 & S2 & S3 &
MK1 & MK2 &
MV & MQ &
CWE & FWE &
QA1 & QA2 &
Mean \\
\midrule

\multirow{8}{*}{Llama-3.1-8B-Instruct}
& Full Attention
& 93.60 & 100.00 & 99.80
& 99.60 & 86.80
& 99.30 & 100.00
& 37.88 & 86.00
& 80.40 & 61.40
& 85.89 \\

& TopK (Oracle)
& 98.40 & 100.00 & 98.60
& 99.40 & 96.00
& 97.40 & 99.75
& 28.26 & 73.40
& 76.80 & 60.80
& 84.44 \\ \cline{2-14}

& StreamingLLM
& 0.80 & 1.20 & 2.00
& 2.00 & 1.20
& 2.50 & 1.75
& 0.72 & 49.07
& 22.20 & 25.40
& 9.89 \\

& SnapKV
& 99.40 & 95.60 & 0.00
& 97.40 & 18.20
& 43.95 & 83.45
& 10.44 & 48.27
& 78.20 & 61.40
& 57.85 \\

& MagicPIG
& 79.60 & 85.60 & 69.20
& 88.00 & 72.40
& 66.25 & 68.10
& 18.52 & \textbf{84.60}
& 70.80 & 61.00
& 69.46 \\

& Loki
& 93.80 & 98.00 & \textbf{99.40}
& 95.00 & 71.20
& 97.90 & 98.15
& \textbf{55.06} & 58.60
& 72.40 & 61.20
& 81.88 \\

& HATA
& \textbf{100.00} & \textbf{100.00} & 99.20
& 99.40 & 96.80
& \textbf{98.60} & 99.80
& 18.68 & 66.93
& \textbf{79.00} & 59.80
& 83.47 \\

& HHR (Ours)
& 99.20 & 99.60 & \textbf{99.40}
& \textbf{99.60} & \textbf{97.40}
& \textbf{98.60} & \textbf{99.90}
& 31.22 & 70.40
& 78.60 & \textbf{61.60}
& \textbf{85.05} \\

\midrule

\multirow{8}{*}{Mistral-7B-Instruct}
& Full Attention
& 100.00 & 99.80 & 96.40
& 92.00 & 84.60
& 96.40 & 89.70
& 73.18 & 82.73
& 60.40 & 40.60
& 83.26 \\

& TopK (Oracle)
& 100.00 & 99.00 & 96.80
& 91.00 & 76.80
& 97.90 & 95.10
& 76.98 & 76.93
& 59.20 & 40.60
& 82.76 \\ \cline{2-14}

& StreamingLLM
& 1.40 & 1.60 & 1.80
& 1.80 & 2.20
& 2.30 & 1.90
& 1.82 & \textbf{83.20}
& 19.40 & 22.20
& 12.69 \\

& SnapKV
& 98.60 & 29.00 & 0.00
& 13.60 & 1.40
& 1.10 & 3.00
& 7.78 & 47.47
& 58.60 & 37.60
& 27.10 \\

& MagicPIG
& 15.40 & 2.60 & 0.00
& 2.00 & 0.80
& 1.55 & 1.25
& 55.26 & 82.27
& 45.60 & 33.40
& 21.83 \\

& Loki
& \textbf{99.80} & 83.00 & 81.00
& 58.80 & 40.20
& \textbf{95.95} & \textbf{91.40}
& \textbf{64.94} & 72.27
& 44.00 & 36.40
& 69.80 \\

& HATA
& \textbf{99.80} & 93.80 & 81.80
& 82.40 & 52.00
& 81.45 & 68.75
& 29.72 & 73.13
& 53.60 & 39.20
& 68.70 \\

& HHR (Ours)
& 99.60 & \textbf{95.80} & \textbf{92.20}
& \textbf{86.60} & \textbf{64.80}
& 87.30 & 83.45
& 53.56 & 75.07
& \textbf{60.00} & \textbf{40.20}
& \textbf{76.23} \\

\midrule

\multirow{8}{*}{Qwen3-4B}
& Full Attention
& 100.00 & 100.00 & 100.00
& 97.00 & 97.40
& 89.65 & 96.75
& 64.82 & 81.40
& 58.40 & 46.00
& 84.67 \\

& TopK (Oracle)
& 100.00 & 99.80 & 99.20
& 96.20 & 95.00
& 77.95 & 79.55
& 60.20 & 61.13
& 48.60 & 43.40
& 78.28 \\ \cline{2-14}

& StreamingLLM
& 0.80 & 1.75 & 0.25
& 2.50 & 2.25
& 1.65 & 0.56
& 0.90 & 1.58
& 18.75 & 20.25
& 4.66 \\

& SnapKV
& \textbf{100.00} & 75.60 & 0.00
& 86.40 & 5.20
& 16.40 & 70.20
& 14.28 & 45.27
& 56.80 & 40.60
& 46.43 \\

& MagicPIG
& 90.00 & 90.60 & 74.80
& 84.80 & 90.60
& 74.35 & 85.00
& 58.30 & \textbf{77.07}
& 51.00 & 43.40
& 74.54 \\

& Loki
& 99.80 & 93.40 & 96.60
& 93.20 & 73.00
& 14.70 & 13.80
& 52.64 & 68.07
& \textbf{64.60} & \textbf{58.20}
& 66.18 \\

& HATA 
& \textbf{100.00} & 98.80 & 98.60
& 95.80 & 94.60
& 77.20 & 78.15
& 60.44 & 63.40
& 55.40 & 43.20
& 78.69 \\

& HHR (Ours)
& \textbf{100.00} & \textbf{99.60} & \textbf{100.00}
& \textbf{97.60} & \textbf{96.20}
& \textbf{94.25} & \textbf{90.40}
& \textbf{63.44} & 64.47
& 58.80 & 45.00
& \textbf{82.71} \\

\bottomrule
\end{tabular}}
\vspace{-6mm}
\end{table*}

\subsection{Ablation Study}
\label{sec:ablation}



\begin{figure}[t]
    \centering
    \begin{minipage}[c]{0.48\linewidth}
        \centering
        \includegraphics[width=\linewidth]{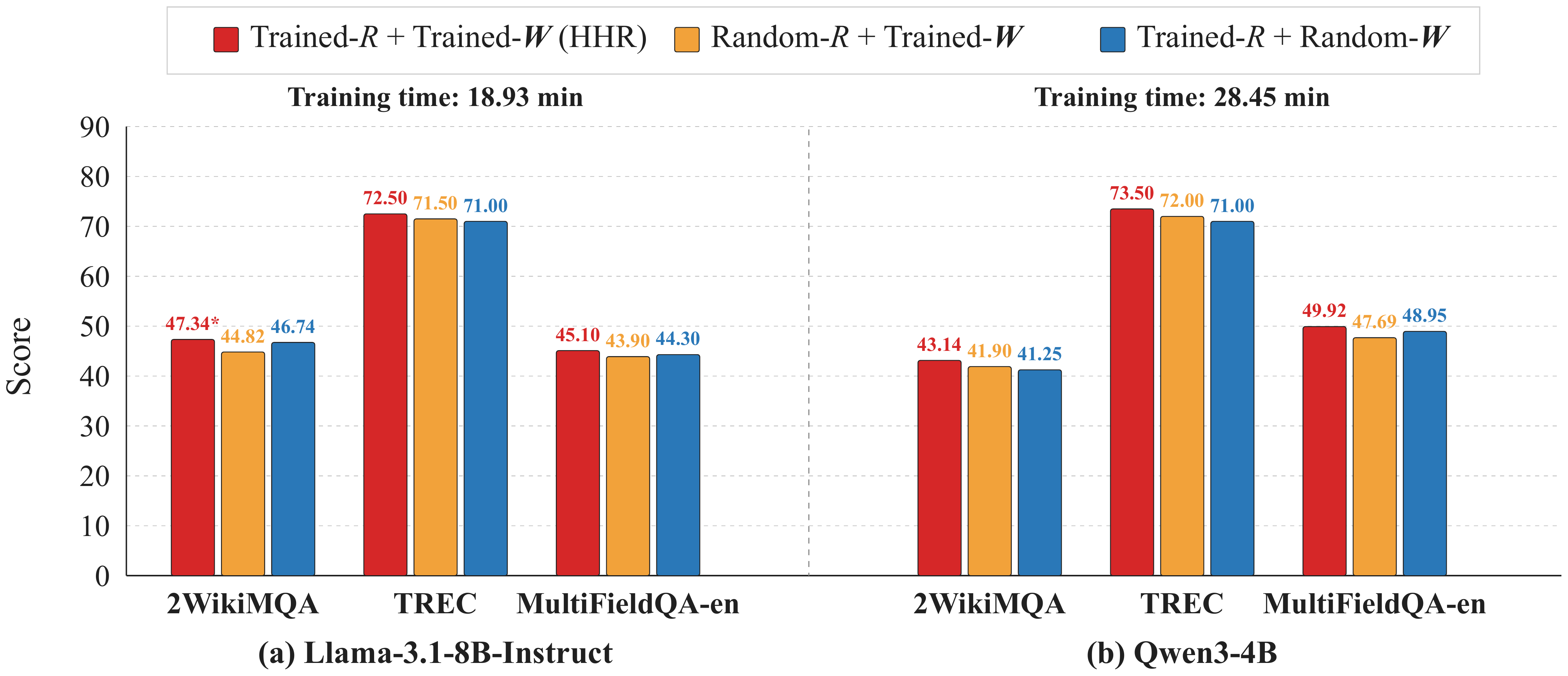}
        \caption{Comparison of the learned $R$ and $\mathbf{W}$. The reported training time denotes the total time required to jointly train $R$ and $\mathbf{W}$.}
        \label{fig:rw_ablation}
    \end{minipage}
    \hfill
    \begin{minipage}[c]{0.48\linewidth}
        \centering
        \includegraphics[width=\linewidth]{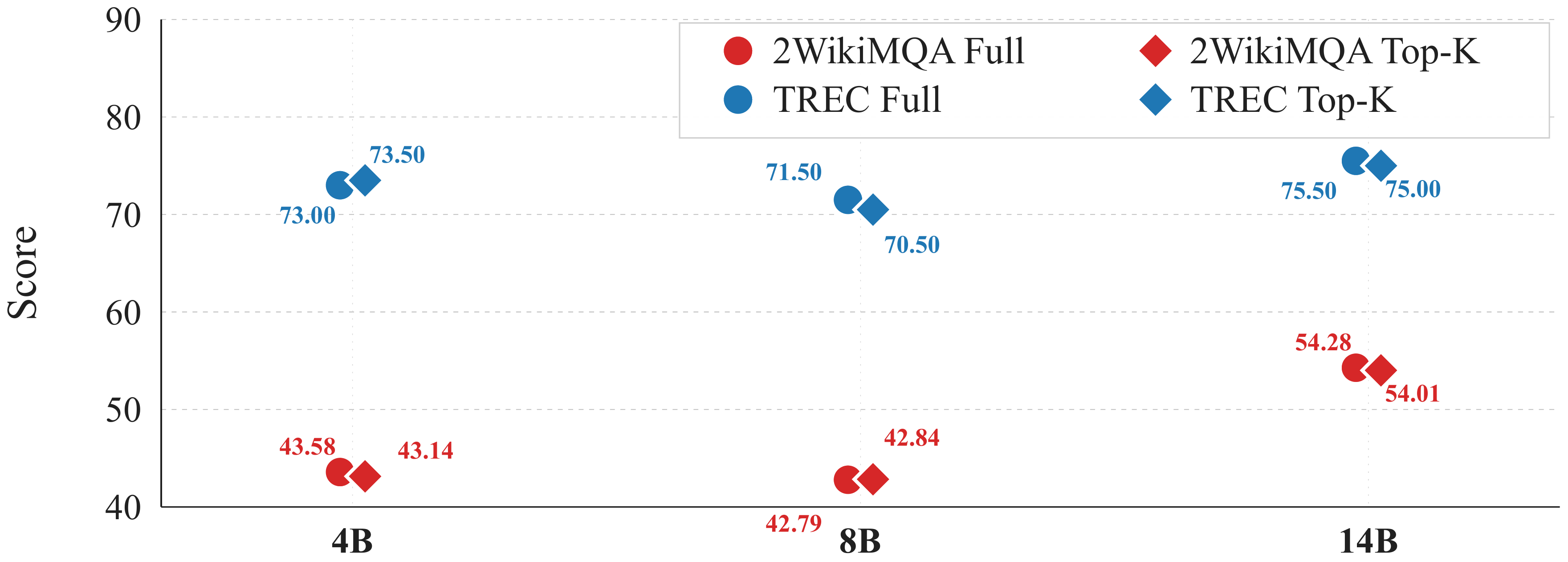}
        \caption{Comparison of Full and HHR Top-K performance across different Qwen3 model sizes.}
        \label{fig:Qwen3}
    \end{minipage}
    \vspace{-4mm}
\end{figure}

\textbf{GKR and LHP.}
To evaluate the roles of $R$ and $\mathbf{W}$ respectively learned in GKR and LHP, we replace them with randomly initialized versions. As shown in Fig.~\ref{fig:rw_ablation}, the datasets include 2WikiMQA~\citep{ho-etal-2020-constructing}, TREC~\citep{LiRo02}, and MultiFieldQA-en~\citep{bai2024longbench}. HHR achieves the best performance across all datasets and backbones. After randomizing $R$, the performance drops by up to 2.52 and 2.23 points on Llama-3.1-8B-Instruct and Qwen3-4B, respectively. After randomizing $\mathbf{W}$, the performance drops by 1.50 and 2.50 points, respectively, indicating that both components play important roles in improving performance. In addition, jointly training $R$ and $\mathbf{W}$ takes \textbf{only 18.93 and 28.45 minutes}, indicating that the proposed method incurs only minimal additional training overhead.

\textbf{Generalization Performance across Different Model Scales.}  
To evaluate the generalization ability of HHR across different model scales, we conduct experiments on Qwen3-4B, Qwen3-8B, and Qwen3-14B, as shown in Fig.~\ref{fig:Qwen3}. Under an extremely sparse Top-K budget of only 2\%, our HHR consistently maintains performance close to Full Attention across all three model scales, validating the stable generalization ability of HHR.

\begin{figure}[!t]
    \centering
    \begin{minipage}[c]{0.48\linewidth}
        \centering
    \includegraphics[width=\linewidth]{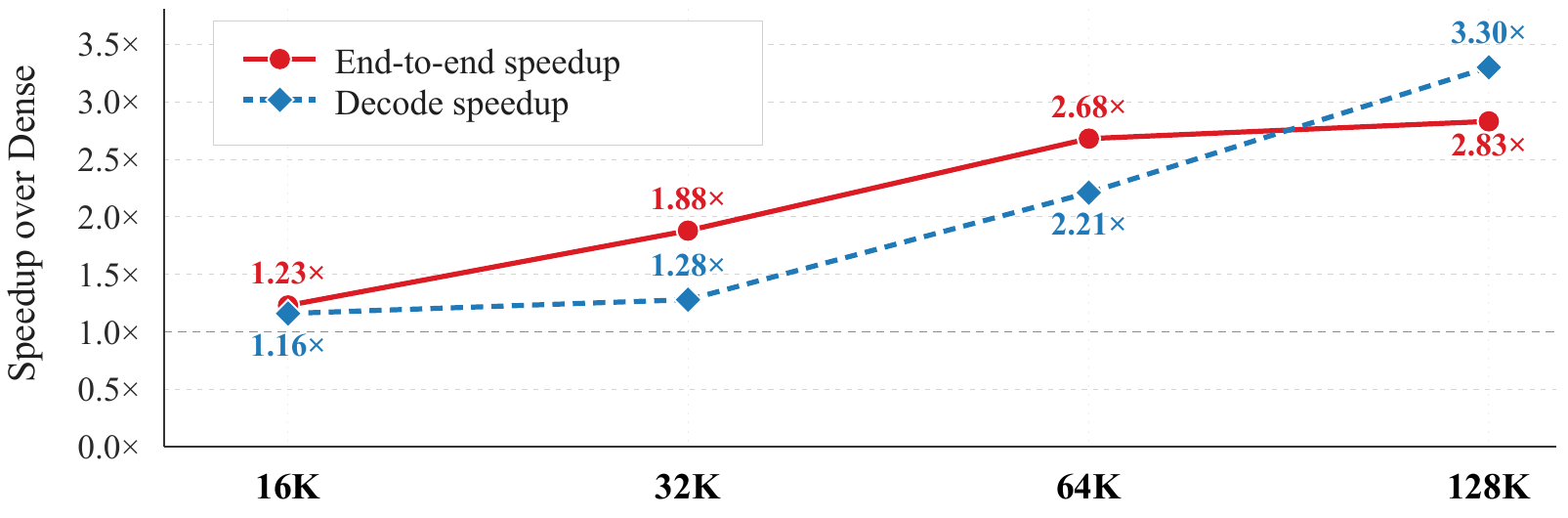}
    \caption{Speedup of HHR over full attention at different context lengths on Llama-3.1-8B-Instruct.}
    \label{fig:context_speedup}
    \end{minipage}
    \hfill
    \begin{minipage}[c]{0.48\linewidth}
        \centering
        \includegraphics[width=\linewidth]{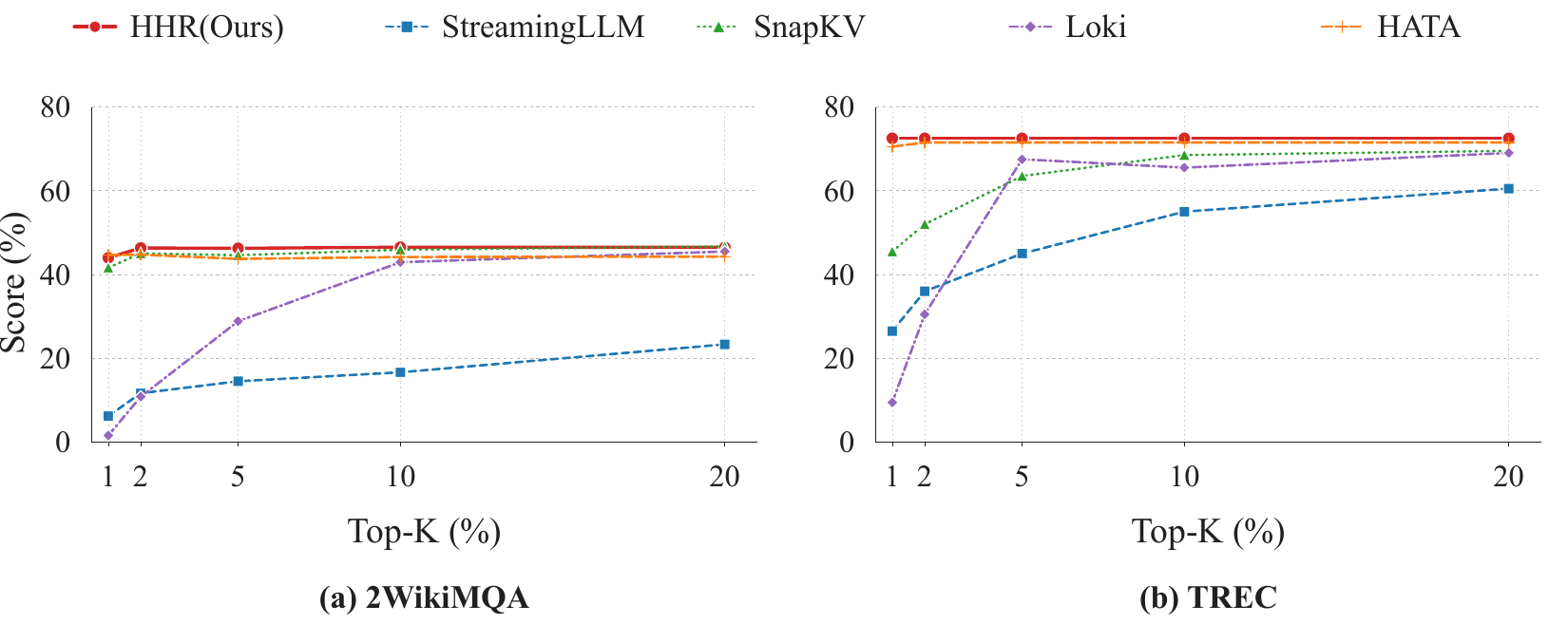}
        \caption{Comparison under different Top-K ratios on Llama-3.1-8B-Instruct.}
        \label{fig:sparsity_ablation}
    \end{minipage}
    \vspace{-5mm}
\end{figure}

\textbf{Context Length.}
We evaluate inference efficiency at context lengths of 16K, 32K, 64K, and 128K. As shown in Fig.~\ref{fig:context_speedup}, decode speedup over dense attention increases from 1.16$\times$ at 16K to 3.30$\times$ at 128K, indicating a greater benefit during decoding as the context grows. End-to-end speedup reaches 2.68$\times$ at 64K and 2.83$\times$ at 128K. These results show that HHR improves efficiency across all evaluated context lengths, with particularly strong gains in long-context decoding.

\textbf{Top-K Ratio.}
We conduct a Top-K ratio ablation on Llama-3.1-8B-Instruct and compare our HHR with other methods. As shown in Fig.~\ref{fig:sparsity_ablation}, HHR stabilizes at around 46.5\% from 2\% onward on 2WikiMQA and consistently achieves 72.50\% on TREC across all Top-K ratios, outperforming the other methods in most settings. More ablation studies are presented in Appendix~\ref{app:More Ablation Study}.
